\documentclass[11pt,a4paper]{article}
\usepackage{authblk}
\usepackage[margin=1in]{geometry}
\usepackage{graphicx}
\usepackage{amsmath}
\usepackage{times}
\usepackage[small,bf]{caption}
\usepackage{abstract}
\usepackage{titlesec}
\usepackage{enumitem}
\usepackage{sectsty}
\usepackage{amsfonts}

\newcommand{\stack}[1]{\!\!\begin{array}{c}\scriptstyle #1\end{array}\!\!}
\newcommand{\brs}[1][-1mm]{\\[#1]\scriptstyle}

\titleformat{\section}[block]{\normalfont\sffamily\large\bfseries\filcenter}{\thesection.}{1em}{}
\titleformat{\subsection}[block]{\normalfont\sffamily\bfseries}{\thesubsection.}{1em}{}
\titleformat{\subsubsection}[block]{\normalfont\sffamily\bfseries}{\thesubsubsection.}{1em}{}
\sectionfont{\centering}
\setlist[enumerate]{topsep=1pt,itemsep=0ex,partopsep=1ex,parsep=1ex}

\title{\vspace{-2.0cm}\sffamily\bfseries\Large Lights out: training RL agents robust to temporary blindness}
\date{\vspace{-7ex}}
\author[1]{Nathan Ordonez} 
\author[2]{Marije Tromp} 
\author[3]{Pau Marquez Julbe}
\author[4]{Wendelin Böhmer}
\affil[1,2,3,4]{\small \emph{Delft University of Technology}}
\affil[1,2,3]{\textit {\{n.a.ordonezcardenas, m.r.tromp, p.marquez\}@student.tudelft.nl}}
\affil[4]{\textit{j.w.bohmer@tudelft.nl}}

\begin{document}
\maketitle


\begin{abstract}\noindent
Agents trained with DQN rely on an observation at each timestep to decide what action to take next. However, in real world applications observations can change or be missing entirely. Examples of this could be a light bulb breaking down, or the wallpaper in a certain room changing. While these situations change the actual observation, the underlying optimal policy does not change. Because of this we want our agent to continue taking actions until it receives a (recognized) observation again. To achieve this we introduce a combination of a neural network architecture that uses hidden representations of the observations and a novel $n$-step loss function. Our implementation is able to withstand location based blindness stretches longer than the ones it was trained on, and therefore shows robustness to temporary blindness.
For access to our implementation, please email Nathan, Marije, or Pau. 

\end{abstract}

\section{Introduction}
\label{sec:introduction}
While modern Reinforcement Learning (RL) agents have successfully learned to complete many different tasks in multiple environments, their robustness to superficial changes in state (changes that shouldn't impact the task at hand) is not a given when using state-of-the-art methods, even though this might be important for safety-critical systems. The name of this problem, as described by a recent survey on RL robustness, is \textit{Observation robustness}\cite{robustrl}. Here, disturbances in the input state are handled by modeling the uncertainty of the environment.

But what should an agent do, when it enters an uncertain state? Specifically, if a state observation has high uncertainty but the agent knows that the underlying environment (specifically, the transition function) has not changed, what should the agent do? A straight-forward to this solution could be that, given the last certain state, an agent could simply remember from the other times it was in that state, what the best sequence of actions should be, and then execute them in an open-loop fashion (meaning, performing actions without paying attention to the states). One could implement this by simply looking at past trajectory recordings, however one would then miss the advantages of using neural networks end-to-end, such as the possibility for better generalization, and the applicability to more complex state spaces. Implementing such an end-to-end neural network and a loss function specialized for this reinforcement learning task is the ultimate goal of this project.

Previous works have explored the open-loop challenge in an offline RL setting. One recent work proposed encoding the input state into a latent representation, and using a recurrent neural network to make future predictions of this latent representation\cite{openlooppaper}. However, the problem they solved was about performing high-frequency actions with access to low-frequency observations, rather than tackling observation robustness. Our approach, and therefore our contributions, are as follows:
\begin{enumerate}
    \item We contribute a novel loss function based on DQN that successfully trains an open-loop policy on recorded trajectories using a recurrent neural network, so that it learns to be robust to open-loop situations during evaluations.
    \item To the best of our knowledge, we contribute the first open-loop loss that is applied to the online deep reinforcement learning setting. In doing so, an agent is able to learn an open-loop policy while exploring and learning to perform a novel task (i.e. a new maze).
    \item Finally, experiments are shown to qualitatively understand the strengths, weaknesses and relevant parameters for agents trained using our loss function in a gridworld maze setting.
\end{enumerate}

\noindent
In this project we are interested in how well an agent performs when navigating a maze while being blinded.
This means that the environment includes a mask, which defines the positions at which the agent does not receive a state vector, and is therefore required to perform open loop control.
We then train an open-loop policy on a set number of blind steps, which is a parameter to experiment with. Our research questions are:
\\[2mm]
\textbf{Can a DQN agent learn a policy that is robust to temporarily not having access to the state during evaluation by switching between closed- and open-loop control?}
\\[2mm]
This main question can be divided into multiple sub-questions:
\begin{enumerate}
    \item Can our loss function train both an open-loop and a closed-loop controller to be robust to blind state masks without training on them?
    \item How does length/frequency of blindness during evaluation affect agent's ability to complete its task?
    \item How does the presence or absence of open-loop policy trajectories in the replay buffer affect robustness to temporary blindness?
    \item How many successive steps in the future can we use to train an open-loop policy robust to temporary blindness? 
\end{enumerate}

The paper is organized as follows: In Section \ref{sec:background} and \ref{sec:related-work} we will outline the basic theory on which our architecture is based. Then, the proposed algorithm and training procedure is detailed in Section \ref{sec:methodology}, followed by an explanation of the experiments in Section \ref{sec:experiments}. Finally, our conclusions are presented in Section \ref{sec:conclusions}.


\section{Background}
\label{sec:background}
First of all, we define a fully-observable MDP by the tuple $\langle \mathcal{S}, \mathcal{A}, r, \mathcal{P}, \mu, \gamma \rangle$, where $\mathcal{S}$ denotes the state space, $\mathcal{A}$ the action space, $r: \mathcal{A} \times \mathcal{S} \mapsto \mathbb{R}$ is the reward function, $\mathcal{P}:\mathcal{S} \times \mathcal{A} \times \mathcal{S} \mapsto [0, 1]$ is the state transition function, $\mu$ is the initial state distribution and $\gamma \in [0, 1]$ is the discount factor. Moreover, \(s_t \in \mathcal{S}\) is the state, \(a_t \in \mathcal{A}\) is the action, and \(r_t \in \mathbb{R}\) is the reward at time step \(t\).

\subsection*{Deep Q-Networks}



Deep Q-Networks (DQN), as introduced by \cite{first_dqn}, are a family of deep Q-Learning algorithms where a value function is estimated for every state-action pair in the system. The goal of reinforcement learning is to find a stationary policy $\pi: \mathcal S \times \mathcal A \to [0,1]$
that maximizes the expected discounted sum of future rewards:
\begin{equation}
    \begin{aligned}
\label{state-value function}
    J^\pi = \mathbb{E}\left[\sum_{t=0}^{\infty} \gamma^t r_t
    \Bigg| \stack{s_0 \sim \mu, \;
            a_t \sim \pi(\cdot|s_t)
        \brs r_t = r(s_t, a_t)
        \brs s_{t+1} \sim \mathcal P(\cdot|s_t, a_t)} \right]
    \end{aligned} \,.
\end{equation}




\noindent
Q-learning achieves this by estimating the Q-value of a greedy policy that selects the action that maximizes that Q-value, yielding the Bellman optimality equation \cite{Sutton18}:
\begin{equation}
    \begin{aligned}
\label{q-value}
    Q(s,a)=\mathbb{E}\big[r(s,a) + \gamma\,\operatorname*{max}_{a^{\prime} \in \mathcal A}Q(s^{\prime},a^{\prime})\big|\stack{s' \sim P(s,a)}\big]
    \end{aligned} \,,
    \qquad \forall s \in \mathcal S,
        \forall a \in \mathcal A \,.
\end{equation}
To deal with large state and action spaces, the Q-value function is implemented as a neural network, and its loss function over transitions $(s_t, a_t, r_t, s_{t+1})$ experienced in the environment is defined as follows:
\begin{equation}
\label{dqn_loss}
    L(\theta) = \mathbb E\Big[\big( r_t + \gamma \max_{a' \in \mathcal{A}} Q_{(s_{t+1}, a'; \theta')} - Q_{(s_t,a_t; \theta)} \big)^{2}\, \Big]
\end{equation}

\noindent
Because the target Q-value function at $s_{t+1}$ evolves at the same speed as the regular Q-value function, training curves can be quite unstable and prevent the agent from converging towards an optimal policy. To avoid this, a \textit{soft target update} is used where the target network, with parameters $\theta'$, is a copy of the original Q-value function, which is updated after every gradient upsate step using the rule $\theta' \leftarrow (1-\tau) \theta' + \tau\theta$ where $\tau$ is the soft update parameter. Further instability both early on during training and after having converged to the optimal policy can happen if an agent only learns from the last step it has taken. To mitigate this, a replay buffer can be helpful. Replay buffers are filled with a set number of past recordings of actions, states and rewards. During training, a batch is sampled from the replay buffer and used to perform gradient descent steps based on the Q-value update function.
\\
In this project, both soft target updates and a replay buffer were used as they improved dramatically the performance of our agent.




\subsection*{Closed and Open-Loop}
The version of DQN described in the previous subsection uses what we call a \textit{closed-loop} controller. Closed-loop controllers receive an observation in every state they are in and take actions based on these observations.

A different type of controller is an \textit{open-loop} controller. These controllers receive an observation and then take multiple actions based on that observation.

A situation in which it is useful to train an open-loop controller is when the state-observation distribution can change, but the optimal action sequence to solve a task does not. An example of this could be the wallpaper in a room changing, or the lights in a hallway turning off. Neither of these will influence the optimal policy, if its goal is to go from one side of the room to the other, for example. A closed-loop controller would not generally be expected to perform well in this scenario, as it is essentially an out of distribution situation for its training, whereas an open-loop controller would simply take actions based off of the last in-distribution observation it has recorded until it reaches the next in-distribution observation. Once the agent is in an in-distribution state, it can switch back to a closed-loop controller and continue its path.


\subsection*{Distribution Shift}
\label{sec:distribution_shift}
\textit{Distribution shift} is a phenomenon that occurs in offline and off-policy learning when the policy applies actions that are out-of-distribution, i.e. not seen during training. To solve this issue, different solutions exist \cite{distribution-shift}, \cite{distribution-shift-2}. This can be a potential issue when we only train the open-loop controller on decisions made by the closed-loop controller. In our case, the amount of open-loop transitions that are sampled during training can be controlled by a parameter to check for this, and the difference in trajectory distributions is indicated by the difference in performance when the agent is evaluated with blindness masks.

\section{Related work}
\label{sec:related-work}
The task of dealing with a lack of state observations can be seen as that of building an agent that can deal with closed and open-loop control. 
\cite{closedOpenLoop1996} describes a Q-learning algorithm that works in an environment where sensing comes at a cost by switching between open-loop and closed-loop, and is the earliest mention of open-loop RL we could find. In our problem sensing does not come at a cost, instead not being available at times. This paper also does not use neural networks.

Open-loop control in deep RL was explored by \cite{openlooppaper}, where the agent has access to a state input only every $n$ steps, thus requiring the agent to make action decisions on the basis of the potentially sparse state inputs. Instead of receiving input every $n$ steps our agent will encounter location based blindness. We therefore test how robust it is to not receiving state inputs.

The concept of encoding the state vector and computing predictions using this embedding has contributed to new state of the art in multiple RL tasks with MuZero \cite{muzero}. Here, the authors use embeddings to build a search tree of possible strategies to play games, evaluating their value and play accordingly. In contrast to this, we plan to exploit previous experience to repeat a known desirable previous behavior and increase agent robustness. Additionally, they do not employ the MuZero algorithm on open-loop control tasks.

In \cite{openlooppaper}, state embeddings are also used to make predictions. The goal there is that the hidden embeddings lead to actions with similar value to actions that would have been taken if the agent had access to the ground truth state input.
The task is also different in terms of the frequency at which the agent performs open-loop control. In their work, the agent is mostly open-loop and regularly gets access to the state, however in our case blindness is not assumed to be regular and the agent is expected to be merely robust to occasional loss of access to state observations.

\section{Methodology}
\label{sec:methodology}
Instead of implementing our DQN agent with an open-loop controller from scratch, we used the DQN implementation from the existing codebase cleanrl \cite{huang2022cleanrl} as a starting point. To transition this implementation into a controller that can handle both closed and open-loop we changed the neural network model architecture, the replay buffer, and loss function. Since our replay buffer is a standard trajectory buffer, we will only explain the architecture and loss functions in the following subsections. 

\subsection*{Architecture}
\begin{figure}
    \centering
    \includegraphics[scale=0.2]{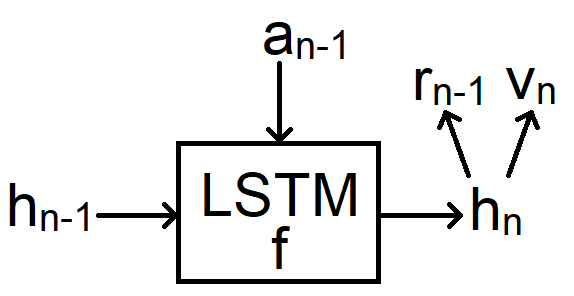}
    \caption{A single step in used architecture.}
    \label{fig:single_block}
\end{figure}

In a closed-loop setting, the controller outputs an action $a_t$ for each state $s_t$. Concretely, with the DQN architecture, it predicts the Q-value of each state-action pair and selects the optimal one.
However, when open-loop control starts, the agent only has access to the first state, and must make an arbitrary number of actions on the basis of this one state.
The architecture we used to learn the open-loop controller is inspired by MuZero \cite{muzero}.
For each open-loop timestep, instead of taking the actual state as the input, our architecture takes a hidden representation of said state.
This hidden representation is not directly regularized, meaning that the neural network learns by itself what information to store in this embedding. This also means that this is a model-based algorithm, in the sense that the neural network can approximate a transition function of the environment.

\begin{figure}[b!]
\begin{minipage}{0.5\textwidth}
    \centering
    \includegraphics[scale=0.2]{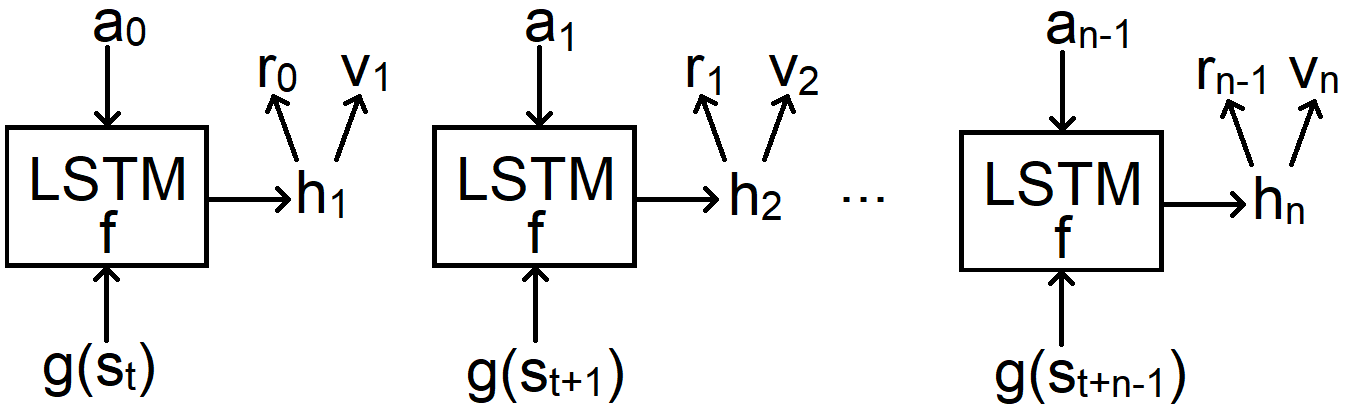}
    \caption{Architecture in closed-loop mode.}
    \label{fig:closed-loop-arch}
\end{minipage}
\begin{minipage}{0.5\textwidth}
    \centering
    \includegraphics[scale=0.2]{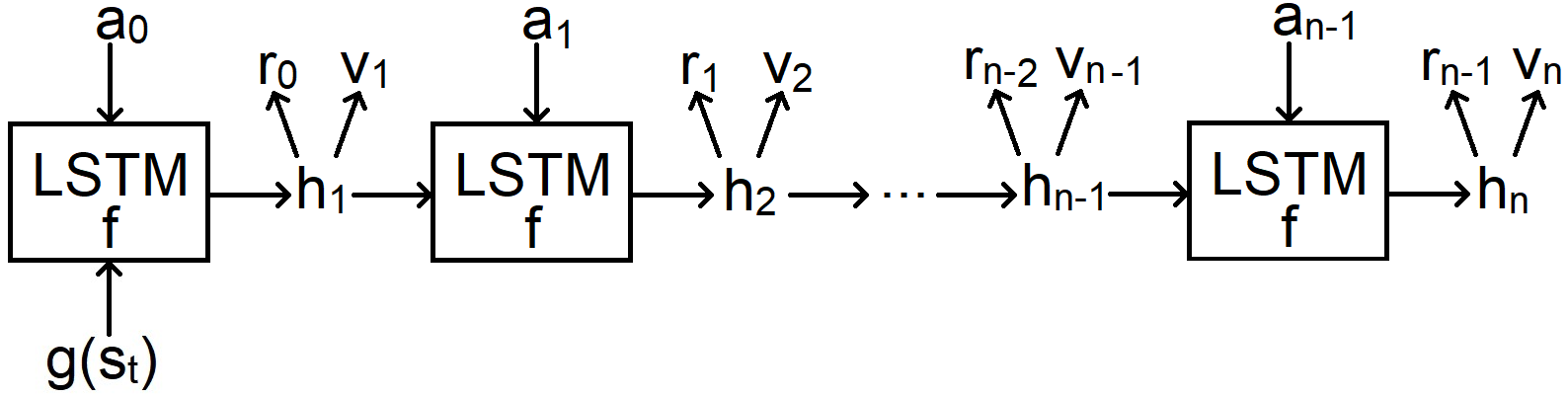}
    \caption{Architecture in open-loop mode.}
    \label{fig:open-loop-arch}
\end{minipage}
\end{figure}

To get the initial hidden state $h_0$, we use an encoder function \(g_\theta\), which takes as input the state \(s_t\) and gives as output the hidden state \(h_0\). We then use this hidden state together with a sequence of actions as input for a function $f_\theta$, which is an Long Short-Term Memory (LSTM) \cite{LSTM}. Using an LSTM allows us to give an action sequence input of variable length, therefore allowing us to use the same neural network for both the closed- and open-loop controller, 
based on Q-values: 
\begin{equation}
    Q(s, a_{0:{n-1}}) \;=\; 
    \sum_{k=0}^{n-1} \gamma^k 
        \underbrace{r_\theta(h_{k+1})}_{r_k}
    +\, \gamma^n \underbrace{v_\theta(h_n)}_{v_n} \,,
    \qquad h_0 \;=\; g_\theta(s) \,,
    \qquad h_{k+1} \;=\; f_\theta(h_{k}, a_k) \,.
\end{equation}
The LSTM $f_\theta$ then returns a new hidden state, which represents the network's interpretation of what the (hidden) state $h_{k+1}$ looks like after taking the action $a_{k}$ from the given (hidden) state $h_k$. From this new hidden state we use a reward function $r_\theta$ and a value function \(v_\theta\) to predict the value $v_{k+1}$ of the new hidden state and the reward $r_k$ of taking the action in the previous hidden state. A single step of this can be seen in Figure \ref{fig:single_block}. Then, based on whether or not there is a new state available, we can use \(g\) to transform that state and use it as input for the next time step, or we can use the output from \(f\) at the previous timestep as the hidden state input.

In Figure \ref{fig:closed-loop-arch}, we show our architecture applied to the closed-loop setting. If we recursively apply function $f$ to the hidden state, we get an open-loop controller as in Figure \ref{fig:open-loop-arch}.


\subsection*{Loss function}
Because of the potential distribution shift when training the open-loop controller on closed-loop trajectories, we can train both policies on trajectories drawn from both policies.
This is done for a certain n-step target loss $n$ during training by becoming blind with probability $p/n$, where $p$ is a parameter to control how often the agent is blind during training. When the agent becomes blind, it then stays blind for $n$ steps. Then, when performing gradient descent we sample $N$-step trajectories $(s_t, a_t, r_t)_{t=\tau}^{\tau+N}$ from the replay buffer of past episodes and apply the following loss to each one:
\begin{equation}
\begin{aligned}
\label{value_loss}\hspace{-2mm}
    \mathcal{L}_{val}(\theta) &\;=\; \sum_{n=1}^N \mathbb E\biggr[ \Big( v_\theta(h_n;s_t, a_{0:n-1}) - y(s_{t+n})\Big)^2 \biggr] \,,
\\ 
    y(s) &\;=\; \max_{a' \in \mathcal A}\Bigl(r_{\theta'}\big(f_{\theta'}(g_{\theta'}(s), a')\big) + \gamma v_{\theta'}\big(f_{\theta'}(g_{\theta'}(s), a')\big)\Bigl) \,.
\end{aligned}
\end{equation}
The goal of this loss function is that value estimations after $n$ open-loop steps need to be as close as possible to closed-loop value estimations. For each time step \(t+n\), the optimal action in time step \(t+n\) is used as target $y(s_{t+n})$. This target is based on single-timestep embeddings from the $f_{\theta'}$ function, using slowly changing target parameters $\theta'$, whereas the value estimation function $v_\theta$ on the left uses $n$-timestep embeddings with differentiated parameters $\theta$, i.e. function $f_\theta$ is applied $n$ times to the state embedding $g_\theta(s_t)$ using the actions from the trajectory buffer that were performed at those timesteps.
\\
To ground the reward head of the network, we compute the following loss, which enforces the predictions of the reward head \(r_\theta\) to represent the environment rewards for each time step in the trajectory:
\begin{equation}
\label{reward_loss}
    \mathcal{L}_{rew}(\theta) \;=\; \sum_{n=1}^N \mathbb E\biggr[(r_\theta(h_n;s_t) - r_{t+n-1})^2\biggr] \,.
\end{equation}

\noindent
The complete loss is:
$
    \mathcal{L}(\theta) = \mathcal{L}_{val}(\theta) + \mathcal{L}_{rew}(\theta) \,.
$



\begin{figure}[b!]
    \centering
    \includegraphics[scale=0.4]{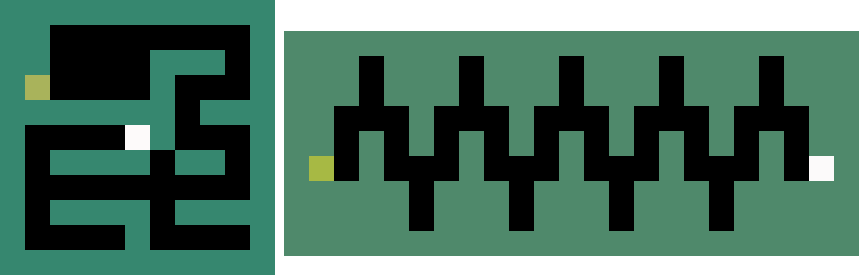}
    \caption{Mazes used in the experiments. In light green the agent's starting point, in white the goal. The left maze is intended to test different kinds of challenges, while the right maze is intended to measure performance for an increasingly long blind area.}
    \label{fig:maze}
\end{figure}

\section{Experiments and Results}
\label{sec:experiments}
\subsection*{Experimental Setup}
To test our controller in different ways, we designed 2 maze environments, as shown in Figure \ref{fig:maze}, and multiple masks for each. In both mazes, the agent spawns on the light green square, and its goal is to arrive at the white square.
To better understand the strengths and weaknesses of our approach, the left maze in Figure \ref{fig:maze} was designed to contain three kinds of challenges; the first part is called the \textit{open room}, where there are many different paths to the exit. The second is the \textit{zigzag path}, chosen because early experimentation indicated it presented a tough challenge for agents. The third is the \textit{forks}, where making a wrong turn could bring you far from the end goal, to punish bad individual decisions. The second maze was designed to empirically check how long of a blindness mask our agents can still solve.
In both mazes, when the agent finds the goal it receives a reward of $+1$, when it walks into a wall, and thus stays in the same spot, it receives a reward of $-0.02$, and when it takes a valid step it receives a reward of $-0.01$.

\begin{figure}
    \centering
    \includegraphics[scale=0.14]{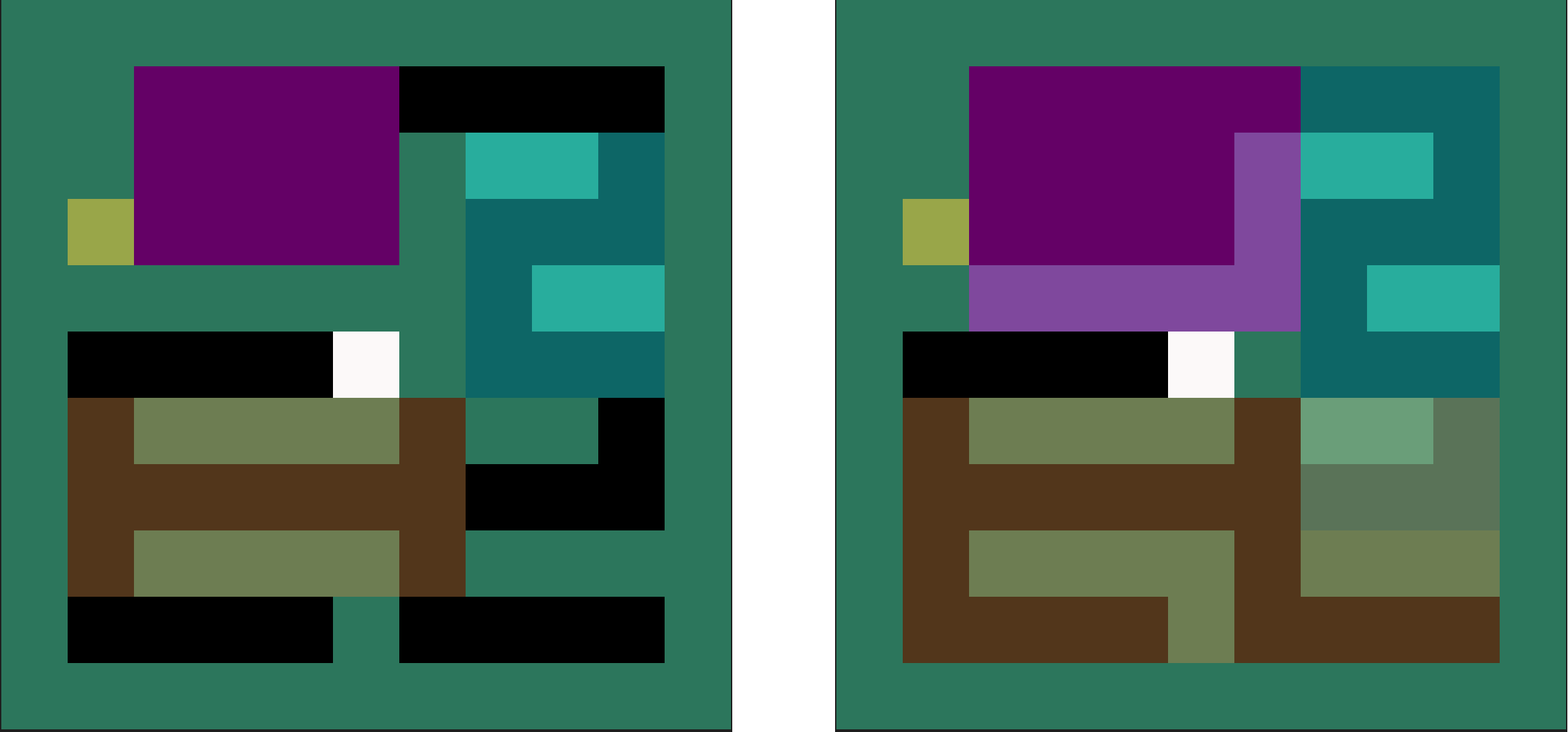}
    \caption{Mazes used in the experiments. The three masks are overlapped in purple, cyan and brown, which mask the open room, zigzag and forks parts, respectively. The light green square on the top left is the starting point of the agent, and the white square is the goal.}
    \label{fig:maze_masks}
\end{figure}
Most importantly, the masks used to determine whether the agent should use open-loop control or not in the first maze, is indicated in Figure \ref{fig:maze_masks}. The left maze shows the masks used together to serve as a benchmark for the agent, so that it faces all three challenges while being blinded. Results for this are in Figure \ref{fig:disjoint-mask}, which shows the performance of the agent for different values of N. The masks on the right were used individually in order to understand which challenges are most difficult for our agents, and the results are shown in the appendix, Figure \ref{fig:threemasks}. For the maze on the left of Figure \ref{fig:maze_masks}, all of the masks were combined to evaluate whether the agents are able to switch between closed and open-loop control. These masks are slightly smaller to facilitate this switch. 

Since we are training a policy to perform on a single maze with a single goal the trained policy would fail to generalize to different mazes or goals. For this reason, the observation given to the policy is the one-hot encoded position of the agent, which is the only dynamic component of the environment. During both training and testing the actions of the agent are restricted to the four vertical and horizontal movements.

To test this project's proposed algorithm, we tried different configurations of hyperparameters on different experimental settings.

\subsection*{Hyperparameters}


During our experiments, after finding hyperparameters that seemed to give good performance, we chose two as being primarily relevant to test in this project.

The first is parameter $p$, which influences the probability that an open-loop trajectory of length $N$ is recorded into the trajectory buffer during training, which is set to $p/N$ to adjust for different values of N.
The second is parameter $N$, which determines the length of the trajectories sampled from the replay buffer during gradient descent, and the maximum length of the $n$-step target losses applied for one neural network update.

\subsection*{Experiments}
To answer our research questions, we tested the performance of our algorithm in 3 different settings, which use different shape blindness masks.
In the areas covered by the masks, the agent will use the open-loop policy to move towards the goal in the optimal amount of steps. 


The first experimental setting serves as an overall benchmark and answers our first sub-question: \textit{Can our loss function train both an open-loop and a closed-loop controller to be robust to blind state masks without training on them?}
For this experiment, we use the left maze in Figure \ref{fig:maze}, combined with all the masks shown in the left maze in Figure \ref{fig:maze_masks}. Since none of these masks overlap, the controller is forced to switch between both open and closed loop control.


The second experimental setting answers our second and third sub-questions: \textit{How does length/frequency of blindness during evaluation affect loss and reward during testing?} and \textit{How does the presence or absence of open-loop policy trajectories in the replay buffer affect robustness to temporary blindness?}. It also attempts to answer our fourth sub-question: \textit{How many successive steps in the future can we use to train an open-loop policy robust to temporary blindness?}. In this experiment we use the right maze in Figure \ref{fig:maze}. For multiple values of $N$ of our proposed loss, we find out the largest number of steps for which it can use the open-loop controller reliably. We do this by starting with the first block of the optimal path being masked, then the first 2, the first 3, etc. until all blocks on the optimal path between the starting position and the goal are covered, or until the agent can no longer reliably reach the goal.

The third experimental setting takes a closer look at the sub-question \textit{Can our loss function train both an open-loop and a closed-loop controller to be robust to blind state masks without training on them?} by using the left maze in Figure \ref{fig:maze}, combined with the masks shown in the right maze in Figure \ref{fig:maze_masks}. We report the relationship between the relevant hyperparameters and each individual mask, to better understand the kinds of challenges that our agents struggle most with. This can be found in the appendix.


\subsection*{Results and Discussion}
\subsubsection*{Switching between closed and open-loop}
\begin{figure}
    \centering
    \includegraphics[scale=0.55]{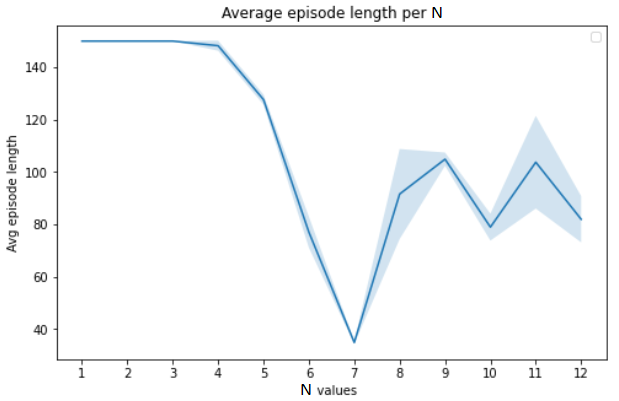}
    \caption{Average episode length on left masks of Figure \ref{fig:maze_masks} over $3$ seeds with different $N$ parameter during training, showing that higher N values allow the agent to solve the maze despite the blindness masks.}
    \label{fig:disjoint-mask}
\end{figure}
In this experiment we trained agents with values of $N$ from 1 to 12 for $60.000$ global steps in the left maze in Figure \ref{fig:maze} with $p$ set to $0.5$. The optimal path for this maze is $ 34$ steps. We evaluate whether the controller is able to switch between closed and open-loop using the left maze in Figure \ref{fig:maze_masks} with epsilon set to $0.05$. The results of this experiment are shown in Figure \ref{fig:disjoint-mask}.

As we can see in the graph, agents that have been trained for $N < 4$ are not able to find the goal within the given timesteps, which might be due to the shortest masked path being 6 steps long. 

What is perhaps more interesting, is that while all of the others were able to reach the goal at some point, only the agent with $N = 7$ was able to do so in (close to) the optimal number of steps. The longest optimal path through any of the masks takes 8 steps. This means that the agent can still generalize at least somewhat to states that it has not been trained on. 

Overall, the Figure shows that the controller is able to successfully switch between open and closed-loop to solve the masked maze, in some cases even performing optimally. 
 However, since we only used 3 different draws to compute the statistics, the results are not statistically reliable enough.

\subsubsection*{Maximum blind length}

In this experiment we trained agents with values 1-25 for $N$ for 45000 global steps in the zigzag maze in Figure \ref{fig:maze} with $p$ $0.0$ and $0.5$. The optimal path for this maze is $40$ steps. As mentioned before, we iteratively test how much blindness the agent can handle by adding a mask to each block on the optimal path one by one. If the agent is able to finish the maze within 150 maximum steps it moves on to the next mask size. To evaluate the actual policy we set epsilon to 0, otherwise the agent would be able to pass through the next mask size on accident. 

\begin{figure}
    \centering
    \includegraphics[scale=0.54]{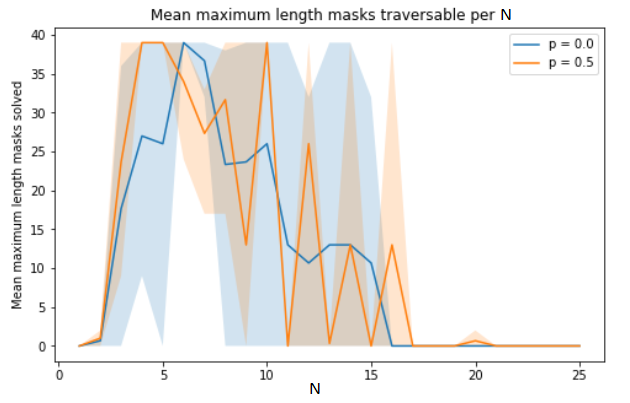}
    \caption{Maximum length of the blindness mask that an agent was able to solve, when trained on different values of N, averaged over 3 seeds. It shows that our method trains an agent robust to blindness lengths multiple times bigger than the $N$ parameter it is trained with, even without sampling any blind trajectories (p=0).}
    \label{fig:max_blind_0_epsilon}
\end{figure}

The results are shown in Figure \ref{fig:max_blind_0_epsilon}, which shows the maximum length the agent was able to pass through for each $N$, calculated from 3 different seeds. The shaded areas are between the minimum blindness length and the maximum blindness length that the agent was able to traverse in these 3 seeds. 

As we can see in the Figure the controller is able to solve masks that are many times the size of $N$. Between $N = 4$ and $N = 10$ the controller is even able to solve all masked blocks for some of the seeds. This means that the controller is able to predict the reward and value relatively accurately, even when it has been blind for much longer than it was trained on. 

\begin{figure}
    \centering
    \includegraphics[scale=0.55]{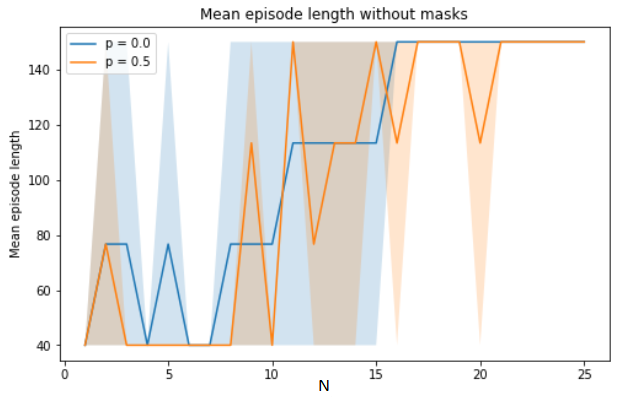}
    \caption{Mean episode length across $3$ different seeds on the zigzag maze with no masks. This figure shows how non-blind performance is affected when varying the $N$ parameter, and explains that the drop-off in performance for higher $N$ values on Figure \ref{fig:max_blind_0_epsilon} is not only because of the blindness.}
    \label{fig:min_episode_no_masks}
\end{figure}
However, for $N$ larger than 10 the performance quickly degrades, and after $N = 16$ it is almost entirely unable to solve any blind steps.
At first glance, it seems like the agent simply loses its ability to be robust to blindness.

However, looking at Figure \ref{fig:min_episode_no_masks}, we notice something else. This graph shows the mean episode length across three seeds for each $N$ for the same zigzag maze without any masks, where epsilon is also set to $0$. We can see that for N larger than 10, the closed-loop controller was unable to solve the problem, which explains why the open-loop controller was not able to solve it either. For this reason, we cannot reach any conclusion about the open-loop controller for $N>15$. A reason why this could be happening is that it needs more training steps to converge, but this has not been tested for. 


This, combined with the fact that we only use 3 different seeds to create these results means that we cannot make any certain conclusions about exactly how the solvable blind period scales with $N$. All we can say is that the performance of the controller is highly dependent on $N$, and that it seems to be able to generalize to blindness lengths far longer than what it was trained with.

The last point of note for this experiment is the 2 different values of $p$. As we can see in the graph, the performance of the open-loop policy does not seem to statistically depend on $p$. One can also see how high the variance is, and in general the results are difficult to interpret with only an average over three seeds.
Since $p$ is influencing the amount of open-loop trajectories that we record during training, we can conclude that there is no distribution shift in this dataset, i.e. the trajectories accessed by both policies during training are similar. This fact could be due to the size of the dataset, where random actions can lead to explore the whole state-space, which would hide the distribution shift.


\subsubsection*{Mask evaluation}
By evaluating the three different masks on the left maze one by one, as depicted on the left of Figure \ref{fig:maze_masks}, we are able to investigate which qualities of the maze give more difficulties to the agent's open-loop controller. Since the experiments showed that hyperparameter $p$ did not have an impact on performance on any of the masks, we will focus on the hyperparameter $N$.
The relationship shows that higher values of $N$ from 1 to 12 generally increase the ability to deal with the three different masks.
Then, comparing the minimum $N$ that achieved the best performance for each mask, we can see that  the \textit{forks, room,} and \textit{zigzag} masks required $N$ values of 3, 4, and 9 to reach optimal performance.
This indicates that the zigzag pattern is especially challenging for the open-loop controller to deal with, and that elongating the $n$-step loss has a major effect on the kinds of challenges the agent is robust to. This could be an indication of something we noticed when looking at episode recordings, that the open-loop controller can get stuck repeating the last action while being blind. More details can be seen in the appendix.


\section{Conclusions}
\label{sec:conclusions}
In this paper we presented a method that uses a recurrent neural network and an $n$-step loss function to learn an open-loop policy using Q-learning. We have showed its high performance and robustness to different kind of mazes in a gridworld by masking the state in different experiment settings. The open-loop policy was successfully able to find the optimal path, even for a large amount of blinded steps.

Overall it is important to keep in mind that the statistics found in this work have high variance due to a small sample of trained neural networks with different seeds.

We have shown that in the small environments we tested, while the trajectory distribution shift between open and closed loop controllers is present as they show differences in performance, training the open loop controller on open-loop sampled trajectories did not have much of an effect on performance. However, since our mazes have a relatively small state space, further experiments in larger mazes should be done in future work to further explore this.


\newpage
\bibliography{references}
\bibliographystyle{unsrt}

\clearpage
\section{Appendix}

\subsection*{A: performance on individual masks}
\begin{figure}[h]
    \centering
    \includegraphics[scale=0.34]{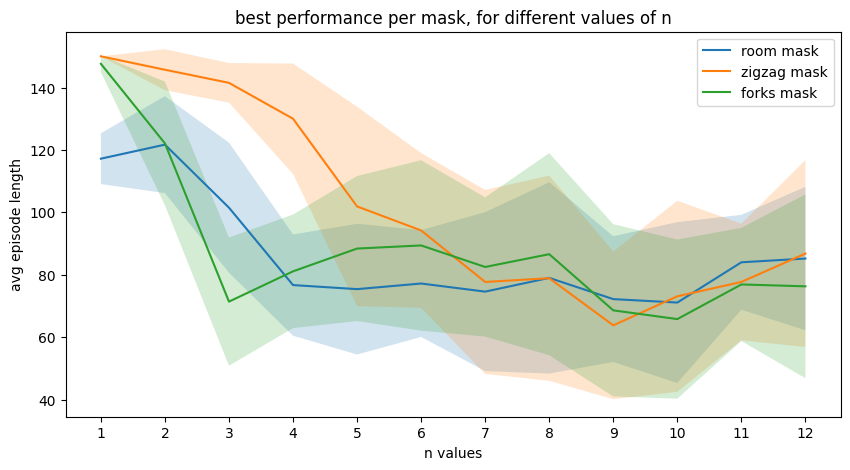}
    \caption{For each value of n, we plot the average and standard deviation of the lowest episode lengths of 9 agents.}
    \label{fig:threemasks}
\end{figure}
Performance on the three masks experimentally significantly depended on parameter n, but not on parameter p, so we plotted the per-mask lowest episode length averaged across the nine values of p that were evaluated between 0 and 0.9. This shows performance for the three qualitatively different challenges of the room, zigzag and fork masks. We can clearly see that the room mask is the easiest to solve with the closed loop (n=1) policy, but that the forks mask can be solved with a lower value of n. The zigzag mask is the one requiring the largest value of n to solve. 150 is the maximum number of steps and indicates that the agent could not reach the goal on time.

\end{document}